%% file: paper.tex
\documentclass[]{bytedance_seed}

\usepackage[toc,page,header]{appendix}

\usepackage{amsmath}
\usepackage{amssymb}
\usepackage{amsfonts}

\usepackage{minitoc}

\title{\raisebox{-0.22\height}{\includegraphics[height=1.15em]{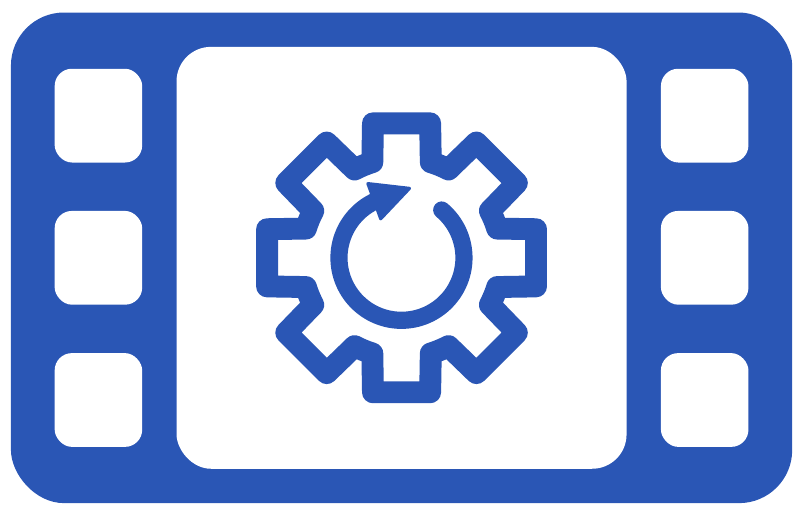}}~\textsc{OpenCoF}: Learning to Reason Through\\Video Generation}

\author[1,2,*]{Xinyan Chen}
\author[3,*]{Ziyu Guo}
\author[1,\dagger]{Renrui Zhang}
\author[1,2]{Dongzhi Jiang}
\author[2]{Hongsheng Li}

\affiliation[1]{ByteDance Seed}
\affiliation[2]{CUHK MMLab}
\affiliation[3]{CUHK IMIXR}

\contribution[*]{Equal contribution}
\contribution[\dagger]{Corresponding author}

\abstract{
Reasoning has become a core capability for large models, especially when reliable decisions require understanding logical consequences. Recent video generation models offer a reasoning path distinct from previous Chain-of-Thought (CoT): reasoning can unfold through temporally connected frames, known as Chain-of-Frame (CoF) reasoning. However, existing video generators are primarily trained on general video corpora, still lacking diverse supervision and dedicated designs for CoF reasoning. To address this gap, we introduce \textsc{OpenCoF}, a framework comprising the \textsc{OpenCoF}-17K dataset, a reasoning video dataset spanning 11 task families, and \textsc{Wan-CoF}, a fine-tuned video model for studying whether diverse temporal supervision improves CoF behavior. Across four video reasoning benchmarks, \textsc{Wan-CoF} achieves considerable gains over the Wan2.2-I2V-A14B baseline. Building on this, we empirically explore more advanced designs for CoF capabilities, i.e., equipping the model with visual and textual reasoning tokens. This mechanism respectively captures low-level visual cues and high-level semantic priors for spatial and temporal reasoning. Through performance comparisons and attention analysis, we examine how these tokens contribute across model depth, denoising steps, space, and time. Our results suggest that stronger video reasoning requires both broad temporal supervision and explicit mechanisms for organizing intermediate reasoning state. We open-source the dataset, model, and code to facilitate future research on reasoning-oriented video generation.
}

\date{\today}
\correspondence{Renrui Zhang at \email{renruizhang@bytedance.com}}

\checkdata[Project Page]{\url{https://opencof.github.io/}}

\begin{document}
\maketitle


\input{sections/introduction}
\input{sections/dataset}

\input{sections/experiments}

\input{sections/exploration}
\input{sections/relatedwork}
\input{sections/conclusion}

\clearpage

\bibliographystyle{plainnat}
\bibliography{main}

\clearpage

\beginappendix

\input{sections/appendix}

\end{document}

%% file: sections/introduction.tex
\section{Introduction}
Enhancing reasoning ability~\cite{Lu2023MathVistaEM, zhang2024mathverse, jiang2025mme, yang2025r1} has emerged as a central objective in the development of large language (LLM) and multimodal (LMM) models. Prior works demonstrate that robust reasoning capabilities are indispensable for reliable decision-making in vision-language contexts. However, mainstream visual Chain-of-Thought (CoT)~\cite{wei2022chain, kojima2022large, meng2025mm} pipelines remain largely anchored to static visual observations. These approaches typically rely on extracting localized evidence~\cite{chen2025mint,shao2024visual}, invoking external tools~\cite{hu2024visual, zheng2025deepeyes, hong2025deepeyesv2,guo2026atlas}, or employing auxiliary image-generation steps~\cite{li2025imagine, xu2025visual}, which limits their ability to intrinsically model dynamic transitions and multi-step visual consequences. 

The rapid advancement of video generation models offers a promising alternative: reasoning via video~\cite{wiedemer2025video, qi2026mme, guo2025video, liu2025can}. Rather than relying solely on textual steps or static visual content, models can reason through the temporal evolution of frames, a process formalized as Chain-of-Frame (CoF) reasoning. Recent studies suggest that video models already exhibit nontrivial world knowledge and early signs of physical and causal understanding, indicating significant potential for CoF~\cite{wiedemer2025video}. Nevertheless, this potential remains far from fully realized. In reasoning-intensive tasks, current video models still struggle with long-term temporal coherence, physical and spatial consistency, and logical continuity~\cite{guo2025video}. Ultimately, high visual realism does not guarantee reliable reasoning.


\begin{figure}[t]
    \centering
    \includegraphics[width=0.9\linewidth]{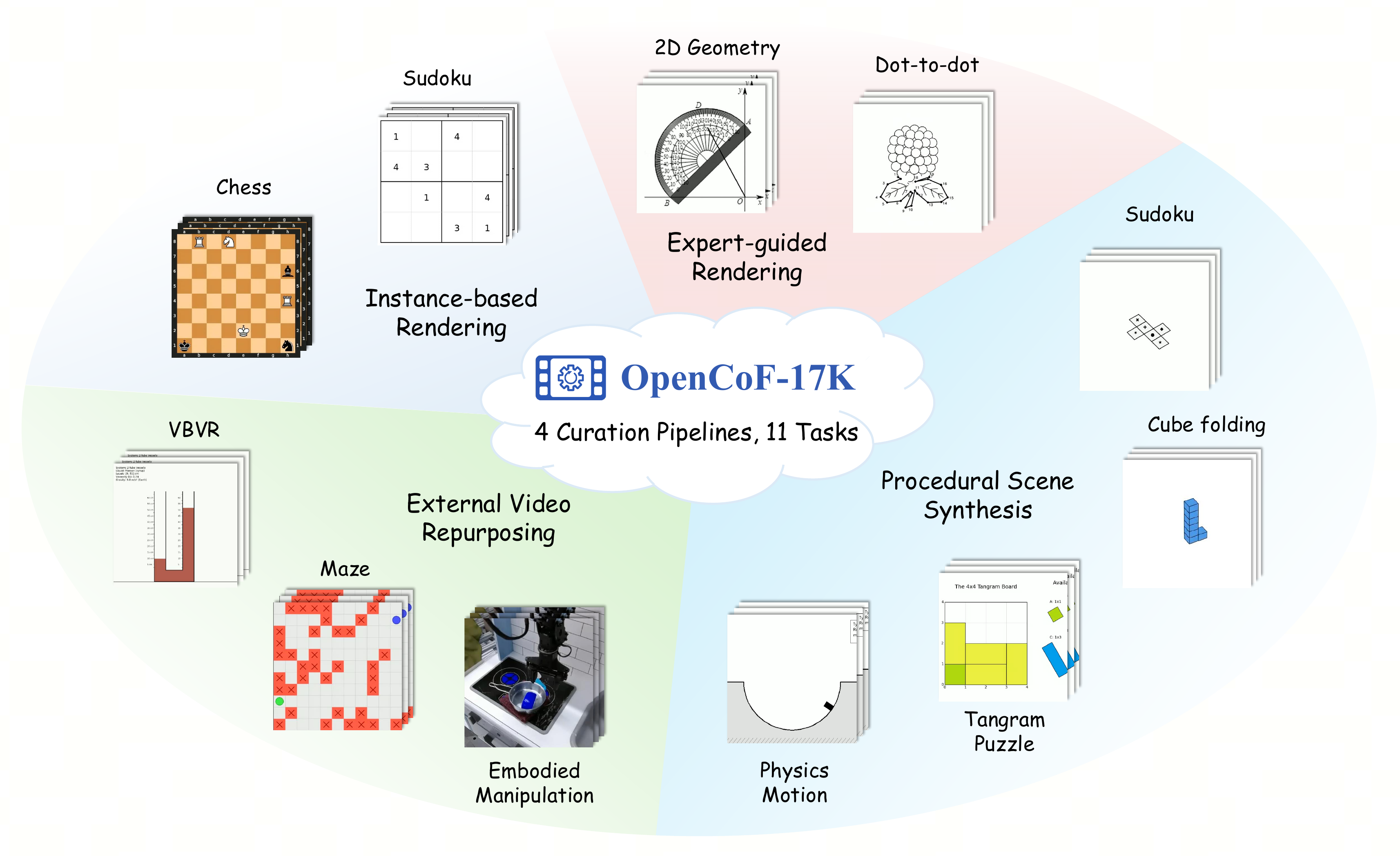}
    \caption{\textbf{Overview of \textsc{OpenCoF}.} We construct the \textsc{OpenCoF}-17K dataset, comprising 17{,}312 videos across 11 tasks, via four complementary pipelines, to provide diverse temporal supervision for Chain-of-Frame (CoF) reasoning.}
    \label{fig:intro_dataset}
\end{figure}


Starting from MME-CoF~\cite{guo2025video}, recent works have predominantly focused on establishing benchmarks to delineate and evaluate the reasoning scope of video generation models, including RULER-Bench~\cite{he2025ruler}, Gen-ViRe~\cite{liu2025can}, VIPER~\cite{li2025viper}, and others~\cite{luo2025v,tong2025thinking}. While these efforts are instrumental in defining the problem space, research dedicated to actively enhancing these reasoning capabilities remains noticeably sparse. A few preliminary studies have explored tailored training and test-time scaling strategies, though they often concentrate on specific task domains, such as maze solving in VR-Bench~\cite{yang2025reasoning} and Thinking in Frames~\cite{li2026thinking}, or custom cognitive-task suites like VBVR~\cite{wang2026very}. Despite these initial advancements, current approaches exhibit two limitations: (1) Their training and evaluation protocols are localized to specific domains or intrinsically coupled to their own curated suites, leaving true generalizability to independent, externally curated reasoning benchmarks underexplored; and (2) their primary emphasis has been on data scaling and inference strategies, leaving internal video model architectures explicitly tailored for reasoning largely uninvestigated. Consequently, there remains a fundamental gap in understanding how to structurally enhance video reasoning capabilities.

To bridge this gap, we introduce \textsc{OpenCoF}, a framework dedicated to advancing video reasoning. As the foundation of this project, we construct the \textsc{OpenCoF}-17K dataset, a diverse collection comprising 17,312 videos across 11 task families. Its curation pipeline integrates four distinct approaches: instance-based rendering, expert-guided rendering, procedural scene synthesis, and the repurposing of existing videos, as illustrated in \Cref{fig:intro_dataset}. Leveraging this dataset, we investigate the enhancement of video reasoning capabilities within a robust open-source video generation model, Wan2.2-I2V-A14B~\cite{wan2025}. First, we fine-tune it on \textsc{OpenCoF}-17K without incorporating any reasoning-specific techniques, yielding \textsc{Wan-CoF}. This data-centric phase aims to verify whether diverse temporal supervision alone can effectively elicit CoF behavior. The subsequent evaluations across four external video-reasoning benchmarks demonstrate substantial gains over the base model.

\begin{figure}[t]
    \centering
    \includegraphics[width=\linewidth]{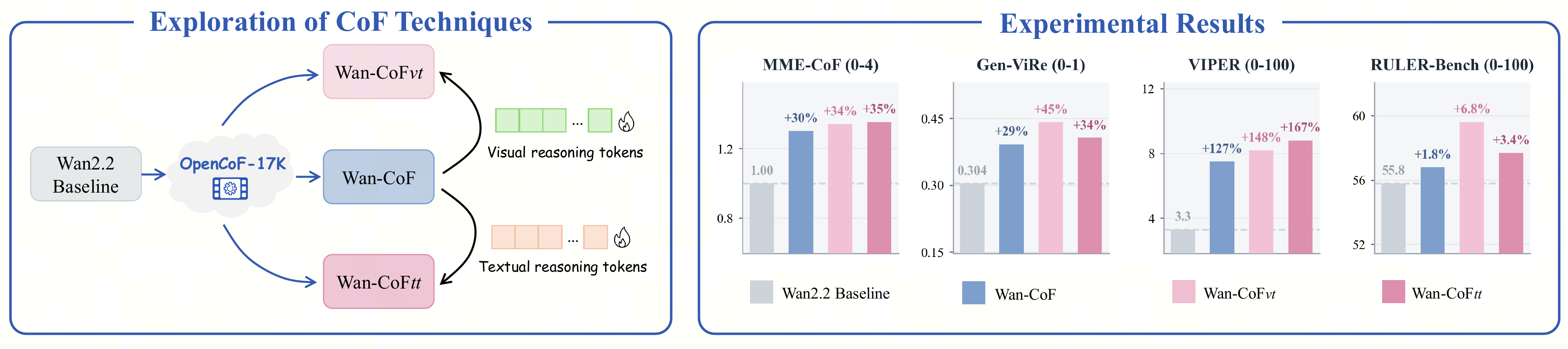}
    \caption{\textbf{Exploration of CoF techniques.} Fine-tuning Wan2.2-I2V-A14B on \textsc{OpenCoF}-17K yields \textsc{Wan-CoF}, which achieves substantial gains over the Wan2.2 baseline across four external video-reasoning benchmarks (MME-CoF, Gen-ViRe, VIPER, RULER-Bench). We further explore Visual Reasoning Tokens (\textit{vt}) and Textual Reasoning Tokens (\textit{tt}) as complementary reasoning-token designs, yielding \textsc{Wan-CoF}$_{{vt}}$ and \textsc{Wan-CoF}$_{{tt}}$, which improve further while exhibiting different strengths across benchmarks.}
    \label{fig:intro_exploration}
\end{figure}

Beyond this data-centric stage, we further explore how dedicated reasoning techniques can help a video generator organize the intermediate reasoning state required by CoF. In current DiT-based video generators, such reasoning remains implicit in generation-oriented visual latents and text conditions. We therefore explore two complementary reasoning-token designs that operate at different levels of the model: Visual Reasoning Tokens (\textit{vt}), inserted into the visual latent sequence to capture low-level visual cues, and Textual Reasoning Tokens (\textit{tt}), appended to the text-conditioning sequence to provide high-level semantic priors. These two variants further improve \textsc{Wan-CoF} on external benchmarks while exhibiting different strengths across task dimensions, as shown in \Cref{fig:intro_exploration}. Through empirical performance evaluations and attention analysis, we examine how these token types facilitate reasoning across model depth, denoising steps, spatial dimensions, and time.

Our contributions are summarized as follows:
\begin{itemize}
    \item We introduce the \textsc{OpenCoF}-17K dataset, comprising 17K videos across 11 task families with a scalable four-source curation pipeline.
    \item We develop \textsc{Wan-CoF} by fine-tuning Wan2.2-I2V-A14B on \textsc{OpenCoF}-17K, demonstrating substantial gains over the base model across four external video-reasoning benchmarks.
    \item We propose complementary reasoning-token mechanisms (\textit{vt} and \textit{tt}) and study their role in CoF reasoning through performance and attention analyses.
\end{itemize}

%% file: sections/dataset.tex
\section{\textsc{OpenCoF}-17K}
Current video generators lack dedicated temporal supervision for reasoning tasks, making diverse reasoning-centric data essential for robust CoF capabilities. To meet this need, we construct \textsc{OpenCoF}-17K, a video reasoning dataset spanning 11 task families. We first summarize its scale, format, and statistics in \Cref{sec:dataset_statistics}, then detail the four-part curation pipeline in \Cref{sec:dataset_curation}.

\subsection{Dataset Statistics}
\label{sec:dataset_statistics}
\begin{figure}[t]
    \centering
    \begin{minipage}[t]{0.50\textwidth}
        \centering
        \caption{Composition of the \textsc{OpenCoF}-17K dataset.}
        \includegraphics[width=0.97\linewidth]{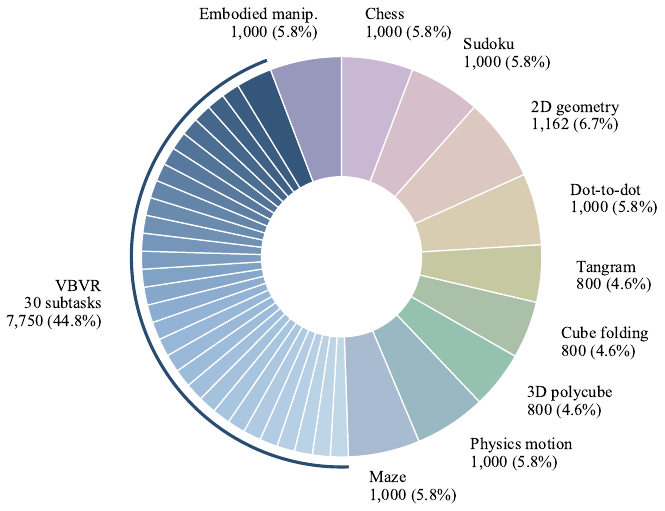}
        \label{fig:dataset_pie}
    \end{minipage}\hfill
    \begin{minipage}[t]{0.49\textwidth}
        \centering
        \captionof{table}{Per-task statistics of \textsc{OpenCoF}-17K. \texttt{Avg prompt} denotes the mean prompt length in words.}
        \label{tab:dataset_statistics}
        \renewcommand{\arraystretch}{0.95}%
        \setlength{\tabcolsep}{4pt}%
        \resizebox{\linewidth}{!}{%
        \begin{tabular}{lcc}
          \toprule
          Task & Num & Avg prompt \\
          \midrule
          Chess                  & 1{,}000  & 42.0  \\
          Sudoku                 & 1{,}000  & 52.0  \\
          2D geometry            & 1{,}162  & 23.4  \\
          Dot-to-dot             & 1{,}000  & 31.0  \\
          Tangram puzzle         & 800  & 13.5  \\
          Cube folding           & 800  & 17.7  \\
          3D polycube rotation   & 800  & 25.3  \\
          Physics motion         & 1{,}000  & 45.9  \\
          Maze                   & 1{,}000  & 179.4 \\
          Embodied manipulation  & 1{,}000  & 10.2  \\
          VBVR (30 subtasks)     & 250 $\times$ 29 $+$ 500  & 62.9  \\
          \midrule
          \textbf{Total}         & \textbf{17{,}312} & \textbf{53.2} \\
          \bottomrule
        \end{tabular}
        }
    \end{minipage}
\end{figure}

\begin{figure}[p]
    \centering
    \includegraphics[width=0.9\linewidth]{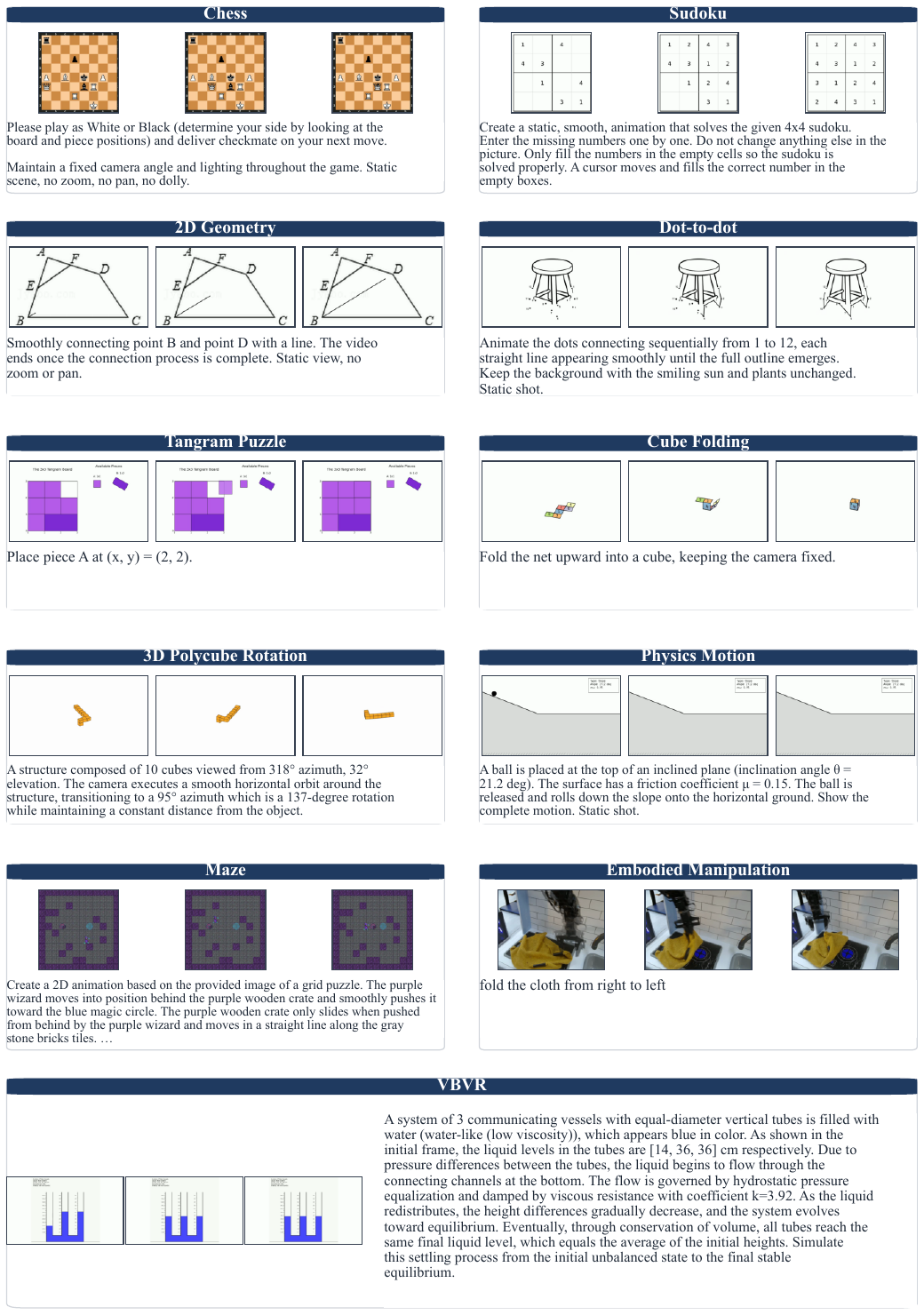}
    \caption{Representative examples from \textsc{OpenCoF}-17K.}
    \label{fig:dataset_examples}
\end{figure}

\begin{figure}[p]
    \centering
    \includegraphics[width=0.9\linewidth]{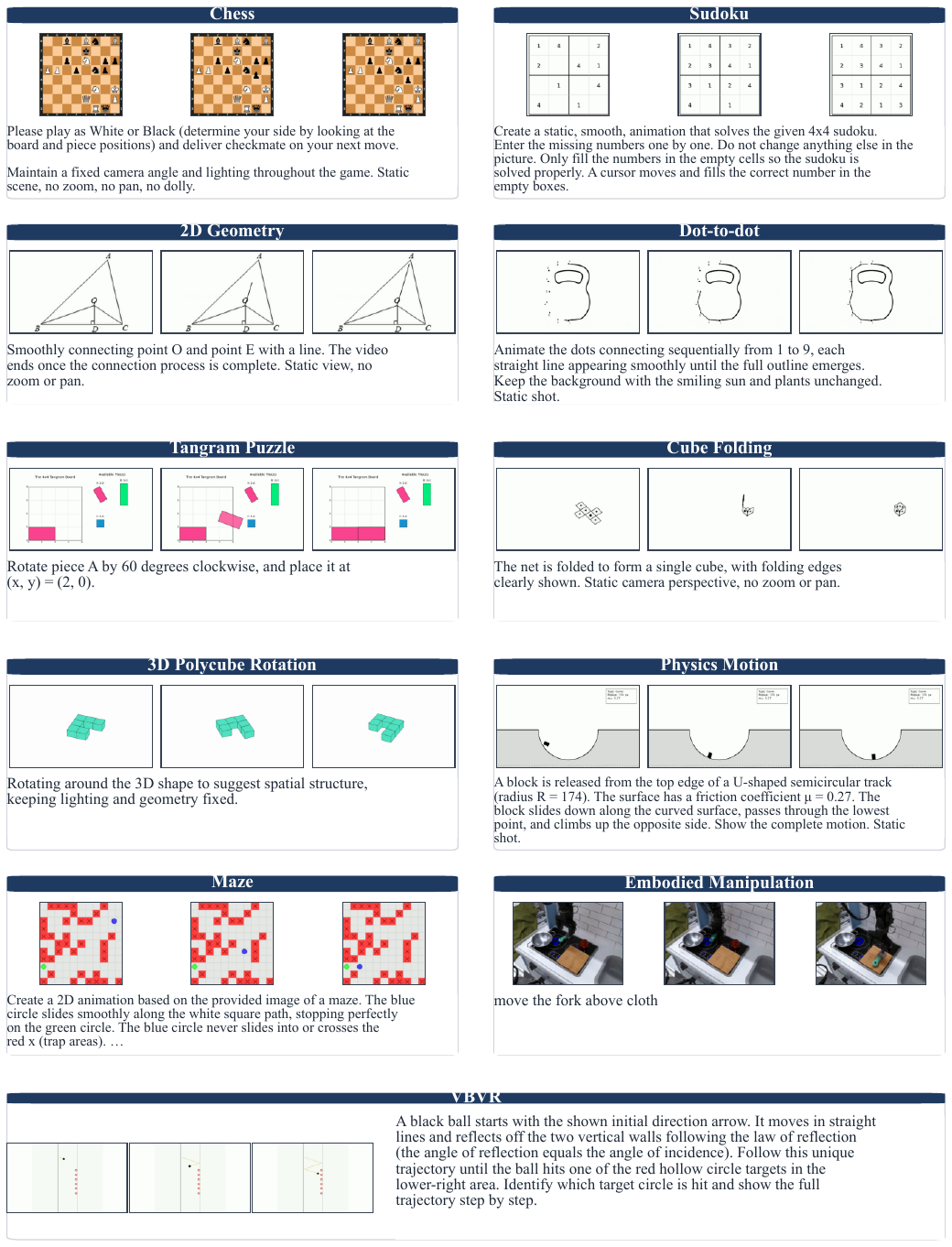}
    \caption{Representative examples from \textsc{OpenCoF}-17K.}
    \label{fig:dataset_examples_v2}
\end{figure}

The \textsc{OpenCoF}-17K dataset comprises 17{,}312 samples across 11 task families. Formulated for conditional generation, each data instance pairs an initial conditioning image and a textual prompt with a target reasoning video. To ensure optimization consistency across diverse data sources, all videos are standardized to a resolution of 480p, a frame rate of 15 fps, and a temporal length of 81 frames. \Cref{tab:dataset_statistics} reports the per-task statistics of \textsc{OpenCoF}-17K. In total the dataset comprises 17{,}312 videos across 11 task families, with an average prompt length of 53.2 words. The VBVR family aggregates 30 subtasks: 29 of them contribute 250 demonstration videos each and the remaining one contributes 500, summing to 7{,}750 videos. 
\Cref{fig:dataset_examples,fig:dataset_examples_v2} show representative cases for each of the 11 task families in the \textsc{OpenCoF}-17K dataset. For every case we display three uniformly sampled frames from the demonstration video together with the original text prompt. 
We next describe how these data are collected and rendered into a unified video-reasoning format.

\subsection{Dataset Curation}
\label{sec:dataset_curation}
\begin{figure}[t]
    \centering
    \includegraphics[width=1\linewidth]{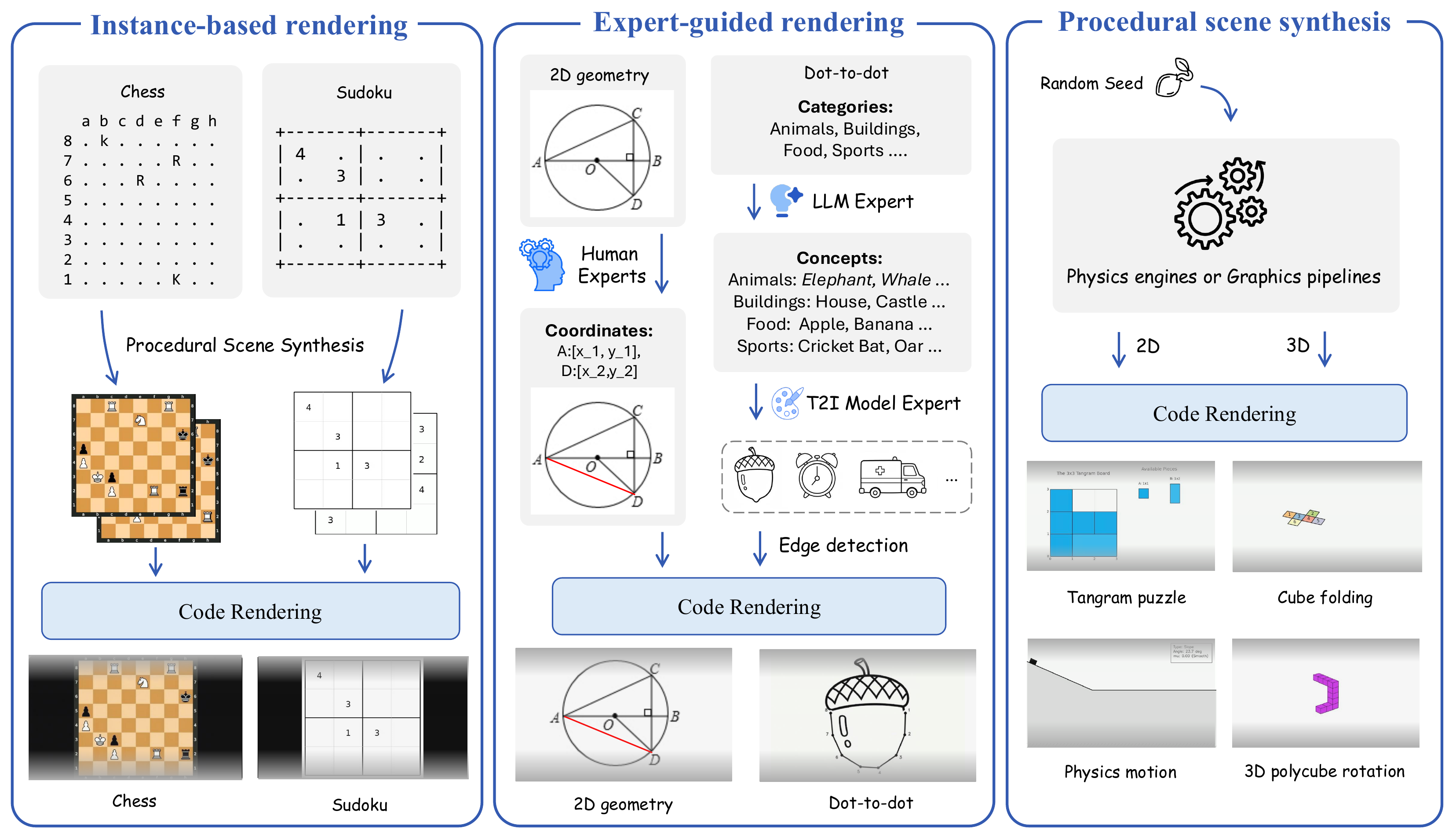}
    \caption{\textbf{Overview of the first three curation pipelines in \textsc{OpenCoF}-17K.} Instance-based rendering converts structured assets into videos through code rendering. Expert-guided rendering uses human, LLM, or T2I-model guidance to define structures before rendering. Procedural scene synthesis uses physics engines or graphics pipelines to instantiate scenes.}
    \label{fig:curation_pipeline}
\end{figure}

To obtain the coverage summarized above, our curation protocol comprises four pipelines: (1)~Instance-based rendering, (2)~Expert-guided rendering, (3)~Procedural scene synthesis, and (4)~External video repurposing. As illustrated in \Cref{fig:curation_pipeline}, the first three pipelines focus on generative construction. To complement these efforts, the fourth pipeline strategically repurposes high-quality demonstrations from existing external datasets. Collectively, these approaches integrate multifaceted data into a unified video-reasoning format, establishing a scalable foundation for future expansion. In contrast to prior efforts like VBVR and VR-Bench that predominantly rely on procedural scene synthesis, our dataset curation pipeline combines multiple complementary construction strategies.

\subsubsection{Instance-based Rendering}
Within this pipeline, we synthesize video data from pre-existing structured assets, such as puzzle states or board configurations. Individual instances are sampled from these assets and sequentially translated into video formats via deterministic code rendering.

\noindent \textbf{Chess.}
This task aims to bolster the reasoning proficiency of video models within highly structured, rule-based environments. We sample instances from the Mate in One (Chess)\footnote{\url{https://www.kaggle.com/datasets/ancientaxe/mate-in-one-chess}} dataset and procedurally render the corresponding video sequences. The models are tasked with identifying and visually executing the singular legal move that results in a checkmate.

\noindent \textbf{Sudoku.}
To inject algorithmic and computational reasoning supervision, we leverage the training split of the \texttt{Sudoku-Easy}\footnote{\url{https://huggingface.co/datasets/LangAGI-Lab/Sudoku-Easy}} dataset, augmenting it with procedurally generated puzzles of equivalent difficulty. Here, video models are required to resolve $4 \times 4$ Sudoku puzzles through a rigorous, step-by-step deductive process.

\subsubsection{Expert-guided Rendering}
Under this protocol, we leverage human expertise, LLM-generated concepts, or outputs from text-to-image (T2I) models to establish structures and annotations. Guided by these representations, we extract or render intermediate structures and convert them into sequential videos with code.

\noindent \textbf{2D geometry.}
Targeting geometric reasoning, we sample 2D geometry images from Geo170K~\cite{gao2023g} that require auxiliary-line construction. Expert annotators define ground-truth coordinates for these segments, after which we render videos showing the step-by-step drawing process.

\noindent \textbf{Dot-to-dot.}
The dot-to-dot task can strengthen the model's understanding of abstract graphics and its ability to follow strict sequential rules. We predefine various categories and prompt LLMs~\cite{comanici2025gemini} to generate specific concepts. Subsequently, we synthesize minimalist line-drawing images using T2I models~\cite{sharon2025nanobanana}, which are then converted into sequential dot-to-dot animation videos via classical computer vision routines.

\subsubsection{Procedural Scene Synthesis}
This pipeline synthesizes video data entirely from scratch. We utilize random seeds to instantiate foundational scene parameters, while physics engines or graphics pipelines subsequently render the unfolding dynamics into continuous video streams.

\noindent \textbf{Tangram puzzle.}
Designed to cultivate spatial awareness and shape manipulation capabilities, this task requires models to systematically rearrange tangram pieces into a target configuration. We procedurally generate puzzle pieces with randomized shapes and initial placements, directly rendering the sequential assembly process into visual formats.

\noindent \textbf{Cube folding.}
Targeting 2D-to-3D spatial reasoning, we procedurally generate cube-folding sequences. Given a flat 2D cube net, the models must predict the correct folding sequence to form a complete 3D cube. The initial nets are randomized across varying topologies, viewing angles, and surface patterns, with the folding dynamics programmatically animated using Matplotlib 3D.

\noindent \textbf{3D polycube rotation.}
To inject geometric motion control supervision across diverse perspectives, we formulate a 3D polycube rotation task. Here, polycubes are instantiated with randomized complex geometries and viewpoints, and are subsequently rotated across varying axes and angles.

\noindent \textbf{Physics motion.}
Physics-based motion reasoning serves as a capability for grounding video models in physical reality. We simulate scenarios featuring balls and blocks traversing flat or curved tracks under varying friction coefficients (e.g., smooth versus rough surfaces). Models are tasked with forecasting the plausible dynamic trajectories of these objects strictly from their initial states.

\subsubsection{External Video Repurposing}
Complementing the three active-construction pipelines visualized in \Cref{fig:curation_pipeline}, we strategically leverage high-quality demonstrations from existing video datasets. These resources are repurposed by standardizing their spatial resolution, frame rate, and temporal length to seamlessly align with our unified training suite.

\noindent \textbf{Maze.} We sample maze navigation data from the training split of the VR-Bench~\cite{yang2025reasoning} dataset. Since some instances in this dataset follow similar distributions, we sample a subset while preserving coverage across subtasks and difficulty levels.

\noindent \textbf{VBVR.} Because some VBVR subtasks are overly simple or target similar capabilities, we select 30 representative subtasks from the broader VBVR~\cite{wang2026very} dataset and randomly sample demonstration videos from each.

\noindent \textbf{Embodied manipulation.}
The data for the embodied manipulation task are sampled from BridgeData V2~\cite{walke2023bridgedata}. This task requires video models to comprehend spatial relationships and perform controlled manipulations.

%% file: sections/experiments.tex
\begin{table*}[t]
  \centering
  \small
  \caption{\textbf{Results on MME-CoF~\citep{guo2025video}.} The scores are reported on a 0--4 scale.}
  \label{tab:main_results}
  \begin{tabular}{l|c|ccccc}
    \toprule
    \multirow{2}{*}{Model} & \multirow{2}{*}{Overall} & Instruction & Temporal & Visual & Content & Focus \\
    & & Alignment & Consistency & Stability & Fidelity & Relevance \\
    \midrule
    \multicolumn{7}{l}{\textit{Closed-source Models}} \\
    Kling-v1~\cite{kuaishou2024kling} & 0.71 & 0.09 & 0.21 & 2.46 & 0.30 & 0.48 \\
    Seedance-1.0-Pro~\cite{gao2025seedance} & 1.48 & 0.39 & 1.71 & 2.05 & 1.22 & 2.04 \\
    Veo-3.0-Fast~\cite{GoogleDeepMind2025Veo3} & 1.52 & 0.73 & 1.41 & 1.83 & 1.25 & 2.36 \\
    Veo-3.0-Preview~\cite{GoogleDeepMind2025Veo3} & 1.53 & 0.70 & 1.46 & 1.91 & 1.24 & 2.36 \\
    Sora-2~\cite{openai2025sora2} & 1.80 & 0.74 & 1.60 & 2.36 & 1.75 & 2.53 \\
    \midrule
    \multicolumn{7}{l}{\textit{Open-source Models}} \\
    Wan2.2-TI2V-5B~\cite{wan2025} & 0.72 & 0.08 & 0.96 & 1.16 & 0.48 & 0.90 \\
    HunyuanVideo-I2V~\cite{kong2024hunyuanvideo} & 0.98 & 0.23 & 1.38 & 1.52 & 0.68 & 1.08 \\
    \midrule
    \multicolumn{7}{l}{\textit{Data-only Fine-tuning}} \\
    Wan2.2-I2V-A14B (Baseline)~\cite{wan2025} & 1.00 & 0.22 & 0.96 & 1.52 & 0.68 & 1.62 \\
    \textbf{\textsc{Wan-CoF}} & 1.30 & 0.31 & 1.29 & 1.99 & 1.00 & 1.89 \\
\textit{\textcolor{blue}{$\Delta$ over the Baseline}}
& \textcolor{blue}{\textit{+0.30}} & \textcolor{blue}{\textit{+0.10}} & \textcolor{blue}{\textit{+0.33}} & \textcolor{blue}{\textit{+0.47}} & \textcolor{blue}{\textit{+0.32}} & \textcolor{blue}{\textit{+0.28}} \\
    \midrule
    \multicolumn{7}{l}{\textit{Exploration of Reasoning Techniques}} \\
    \textbf{\textsc{Wan-CoF}$_{{vt}}$} & 1.34 & 0.26 & 1.38 & 2.33 & 0.99 & 1.76 \\
    \textbf{\textsc{Wan-CoF}$_{{tt}}$} & 1.35 & 0.34 & 1.27 & 2.18 & 1.17 & 1.77 \\
    \bottomrule
  \end{tabular}
\end{table*}

\begin{table*}[t]
  \centering
  \small
  \caption{\textbf{Results on Gen-ViRe~\citep{liu2025can}.} Following the Gen-ViRe protocol, the scores are on a 0--1 scale.}
  \label{tab:genvire_results}
  \resizebox{\linewidth}{!}{
  \begin{tabular}{l|c|cccccc}
    \toprule
    Model & Overall & Abstract & Algo. \& Logi. & Analogy & Perceptual & Planning & Spatio-Temporal \\
    \midrule
    \multicolumn{8}{l}{\textit{Closed-source Models}} \\
    Kling-v1~\cite{kuaishou2024kling} & 0.198 & 0.071 & 0.057 & 0.117 & 0.140 & 0.443 & 0.359 \\
    Seedance-1.0-Pro~\cite{gao2025seedance} & 0.301 & 0.154 & 0.164 & 0.083 & 0.171 & 0.609 & 0.621 \\
    Veo-3.1~\cite{GoogleDeepMind2025Veo3} & 0.486 & 0.440 & 0.451 & 0.367 & 0.386 & 0.722 & 0.550 \\
    Sora-2~\cite{openai2025sora2} & 0.560 & 0.604 & 0.472 & 0.483 & 0.496 & 0.768 & 0.537 \\
    \midrule
    \multicolumn{8}{l}{\textit{Data-only Fine-tuning}} \\
    Wan2.2-I2V-A14B (Baseline)~\cite{wan2025} & 0.304 & 0.220 & 0.359 & 0.083 & 0.190 & 0.519 & 0.452 \\
    \textbf{\textsc{Wan-CoF}} & 0.391 & 0.328 & 0.506 & 0.133 & 0.347 & 0.578 & 0.451 \\
\textit{\textcolor{blue}{$\Delta$ over the Baseline}} & \textcolor{blue}{\textit{+0.087}} & \textcolor{blue}{\textit{+0.107}} & \textcolor{blue}{\textit{+0.147}} & \textcolor{blue}{\textit{+0.050}} & \textcolor{blue}{\textit{+0.157}} & \textcolor{blue}{\textit{+0.059}} & \textcolor{blue}{\textit{-0.001}} \\
    \midrule
    \multicolumn{8}{l}{\textit{Exploration of Reasoning Techniques}} \\
    \textbf{\textsc{Wan-CoF}$_{{vt}}$} & 0.441 & 0.475 & 0.507 & 0.233 & 0.298 & 0.628 & 0.507 \\
    \textbf{\textsc{Wan-CoF}$_{{tt}}$} & 0.406 & 0.424 & 0.474 & 0.167 & 0.306 & 0.565 & 0.499 \\
    \bottomrule
  \end{tabular}
  }
\end{table*}

\section{Training of \textsc{Wan-CoF}}
\label{sec:experiments}

Before introducing explicit reasoning techniques in the models, we first conduct the most direct study: how far can a strong open video generator go when it is trained only with diverse CoF supervision? To answer this, we use Wan2.2-I2V-A14B~\cite{wan2025}, a general-domain pretrained I2V model, and directly fine-tune it on \textsc{OpenCoF}-17K with LoRA~\cite{hu2022lora}. The resulting model, \textsc{Wan-CoF}, marks the data-only stage of our study and serves as a clean reference point for attributing performance gains to the dataset itself. We evaluate \textsc{Wan-CoF} on four video reasoning benchmarks, which allows us to separate the effect of \textsc{OpenCoF}-17K from the reasoning-token designs explored later in \Cref{sec:exploration}. Please refer to \Cref{sec:add_impl} for experimental details.

\begin{table*}[t]
\centering
\small
\caption{\textbf{Results on VIPER~\citep{li2025viper}.} We follow the original VIPER protocol and report POC@1.0.}
\label{tab:viper_results}
\resizebox{\linewidth}{!}{
\begin{tabular}{l|c|cccccc}
    \toprule
    Model & Overall & Temporal & Structural & Symbolic & Spatial & Physics & Planning \\
    \midrule
    \multicolumn{8}{l}{\textit{Closed-source Models}} \\
    Seedance-1.5~\cite{seedance2025seedance} & 9.5 & 5.6 & 5.6 & 0.0 & 5.3 & 6.5 & 33.8 \\
    Veo-3.1~\cite{GoogleDeepMind2025Veo3} & 20.3 & 22.2 & 20.0 & 13.3 & 10.7 & 14.7 & 41.0 \\
    Sora-2~\cite{openai2025sora2} & 23.3 & 10.4 & 42.5 & 58.3 & 4.0 & 9.4 & 15.1 \\
    \midrule
    \multicolumn{8}{l}{\textit{Open-source Models}} \\
    Hunyuan-1.5~\cite{wu2025hunyuanvideo} & 8.1 & 8.4 & 5.0 & 0.0 & 2.7 & 11.1 & 21.1 \\
    \midrule
    \multicolumn{8}{l}{\textit{Data-only Fine-tuning}} \\
    Wan2.2-I2V-A14B (Baseline)~\cite{wan2025} & 3.3 & 5.6 & 6.2 & 0.0 & 0.0 & 2.8 & 5.3 \\
    \textbf{\textsc{Wan-CoF}} & 7.5 & 18.8 & 10.0 & 0.0 & 5.3 & 8.3 & 5.3 \\
\textit{\textcolor{blue}{$\Delta$ over the Baseline}} & \textcolor{blue}{\textit{+4.2}} & \textcolor{blue}{\textit{+13.2}} & \textcolor{blue}{\textit{+3.8}} & \textcolor{blue}{\textit{0.0}} & \textcolor{blue}{\textit{+5.3}} & \textcolor{blue}{\textit{+5.5}} & \textcolor{blue}{\textit{0.0}} \\
    \midrule
    \multicolumn{8}{l}{\textit{Exploration of Reasoning Techniques}} \\
    \textbf{\textsc{Wan-CoF}$_{{vt}}$} & 8.2 & 12.8 & 11.3 & 0.0 & 4.0 & 8.3 & 15.8 \\
    \textbf{\textsc{Wan-CoF}$_{{tt}}$} & 8.8 & 14.8 & 15.0 & 0.0 & 8.0 & 2.8 & 10.5 \\
    \bottomrule
\end{tabular}
}
\end{table*}

\begin{table*}[t]
  \centering
  \small
  \caption{\textbf{Results on RULER-Bench~\citep{he2025ruler}.} We report the filtered image-to-video setting on a 0--100 scale, since our baseline does not support text-to-video generation. IF: Instruction Following, VC: Visual Consistency, VF: Visual Fidelity, RC: Rule Coherence.}
  \label{tab:ruler_results}
  \resizebox{\linewidth}{!}{
  \begin{tabular}{l|c|ccccc|ccccc|ccccc|ccccc}
    \toprule
    \multirow{2}{*}{Model} & \multirow{2}{*}{Overall} & \multicolumn{5}{c|}{Vision} & \multicolumn{5}{c|}{Science} & \multicolumn{5}{c|}{Game} & \multicolumn{5}{c}{Humanity} \\
    \cmidrule{3-22}
     & & IF & VC & VF & RC & Avg. & IF & VC & VF & RC & Avg. & IF & VC & VF & RC & Avg. & IF & VC & VF & RC & Avg. \\
    \midrule
    \multicolumn{22}{l}{\textit{Data-only Fine-tuning}} \\
    Wan2.2-I2V-A14B (Baseline)~\cite{wan2025} & 55.8 & 0.0 & 69.1 & 65.8 & 28.0 & 54.3 & 44.2 & 73.5 & 83.6 & 18.4 & 54.9 & 21.6 & 65.6 & 83.4 & 12.0 & 45.6 & 61.1 & 91.7 & 91.7 & 40.2 & 71.2 \\
    \textbf{\textsc{Wan-CoF}} & 56.8 & 0.0 & 65.1 & 67.0 & 31.7 & 54.6 & 45.6 & 70.8 & 83.1 & 18.9 & 54.6 & 25.0 & 74.4 & 87.0 & 23.9 & 52.6 & 47.2 & 86.1 & 94.4 & 49.4 & 69.3 \\
\textit{\textcolor{blue}{$\Delta$ over the Baseline}} & \textcolor{blue}{\textit{+1.0}} & \textcolor{blue}{\textit{0.0}} & \textcolor{blue}{\textit{-4.0}} & \textcolor{blue}{\textit{+1.3}} & \textcolor{blue}{\textit{+3.6}} & \textcolor{blue}{\textit{+0.3}} & \textcolor{blue}{\textit{+1.4}} & \textcolor{blue}{\textit{-2.7}} & \textcolor{blue}{\textit{-0.5}} & \textcolor{blue}{\textit{+0.5}} & \textcolor{blue}{\textit{-0.3}} & \textcolor{blue}{\textit{+3.4}} & \textcolor{blue}{\textit{+8.8}} & \textcolor{blue}{\textit{+3.6}} & \textcolor{blue}{\textit{+11.9}} & \textcolor{blue}{\textit{+6.9}} & \textcolor{blue}{\textit{-13.9}} & \textcolor{blue}{\textit{-5.6}} & \textcolor{blue}{\textit{+2.8}} & \textcolor{blue}{\textit{+9.3}} & \textcolor{blue}{\textit{-1.9}} \\
    \midrule
    \multicolumn{22}{l}{\textit{Exploration of Reasoning Techniques}} \\
    \textbf{\textsc{Wan-CoF}$_{{vt}}$} & 59.6 & 0.0 & 72.7 & 71.0 & 35.6 & 59.8 & 48.5 & 68.5 & 81.4 & 15.5 & 53.5 & 30.7 & 73.4 & 86.4 & 22.1 & 53.2 & 58.3 & 88.9 & 98.2 & 56.9 & 75.6 \\
    \textbf{\textsc{Wan-CoF}$_{{tt}}$} & 57.7 & 0.0 & 71.2 & 68.1 & 34.2 & 57.8 & 45.9 & 69.8 & 84.6 & 18.9 & 54.8 & 28.6 & 76.5 & 87.2 & 23.1 & 53.8 & 52.8 & 86.1 & 94.9 & 49.4 & 70.8 \\
    \bottomrule
  \end{tabular}
  }
\end{table*}

\subsection{Main Results}
\paragraph{Comparison with the baseline.}
As shown in the Data-only Fine-tuning rows of \Cref{tab:main_results,tab:genvire_results,tab:ruler_results,tab:viper_results}, relative to our baseline, Wan2.2-I2V-A14B, \textsc{Wan-CoF} lifts the headline metric on every benchmark: MME-CoF Overall from 1.00 to 1.30 ($+0.30$), Gen-ViRe average from 0.304 to 0.391 ($+0.087$), VIPER POC from 3.3 to 7.5 ($+4.2$), and RULER-Bench overall average from 55.8 to 56.8 ($+1.0$). The gains are broad-based---every MME-CoF aspect improves, and on Gen-ViRe and VIPER almost every category also gains. The severe non-positive cells sit within evaluation noise or correspond to dimensions where open-source models still score near zero. Moreover, the largest jumps cluster on reasoning-oriented sub-dimensions like Temporal Consistency, Physics, Spatial, Perceptual, and Algorithmic \& Logical, rather than on visual fidelity alone, suggesting that our training strengthens frame-level reasoning rather than acting as a visual-quality regularizer.

\paragraph{Comparison with other models.}
In the Data-only Fine-tuning rows of \Cref{tab:main_results,tab:genvire_results,tab:ruler_results,tab:viper_results}, \textsc{Wan-CoF} surpasses open-source competitors and narrows the gap to the closed-source frontier. On MME-CoF, \textsc{Wan-CoF} exceeds HunyuanVideo-I2V and Wan2.2-TI2V-5B, approaching Seedance-1.0-Pro. On Gen-ViRe, \textsc{Wan-CoF} reaches 0.391 average, outperforming Kling-v1 and Seedance-1.0-Pro and moving closer to Veo-3.1. On VIPER, \textsc{Wan-CoF} attains POC of 7.5, on par with open-source Hunyuan-1.5 and closed-source Seedance-1.5. Under a LoRA fine-tuning budget on a single open backbone, this places \textsc{Wan-CoF} among the strongest open-source I2V models on CoF reasoning.

\subsection{Out-of-distribution Analysis}
A key challenge in evaluating CoF reasoning is ensuring that models generalize beyond their training distributions. While recent works~\citep{yang2025reasoning, wang2026very} evaluate distribution shifts internally within their respective task suites, we explore generalization by transferring our model directly to independent, externally curated benchmarks.
Because these external benchmarks encompass distributions distinct from our training set, testing on them measures performance under a distribution shift. Under this setting, \textsc{Wan-CoF} demonstrates consistent improvements across all four benchmarks. Notably, the per-category gains appear to align with the types of supervision provided by \textsc{OpenCoF}-17K: training on Physics motion transfers to VIPER Physics (+5.5); structured-grid tasks (e.g., Chess, Sudoku, Maze) correlate with Gen-ViRe Algorithmic \& Logical (+0.147) and the RULER-Bench Game split (+6.9); and spatial-shape tasks map to Gen-ViRe Perceptual (+0.157) and VIPER Spatial (+5.3). Together, these transfer signals suggest that \textsc{OpenCoF}-17K helps foster transferable CoF reasoning skills across suite boundaries, rather than overfitting to specific training domains.


%% file: sections/exploration.tex
\section{Exploration of CoF Techniques}
\label{sec:exploration}
The data-only experiment in \Cref{sec:experiments} establishes an encouraging starting point: training on \textsc{OpenCoF}-17K transfers to external reasoning benchmarks even when reasoning techniques are not explicitly added. This result also sharpens the empirical question: if CoF gains can emerge from supervision alone, can dedicated reasoning techniques give the model a more explicit place to organize the intermediate rules, plans, and abstract goals behind those gains? The underlying generator still lacks a dedicated channel for maintaining reasoning state, and must express such information implicitly through pixel-level latents. Since existing video generation models are not originally designed for stepwise reasoning, they offer no specialized structures to host either the low-level reasoning cues operating inside the visual latent space or the high-level reasoning semantics operating at the prompt level. We therefore explore two complementary designs to target these distinct layers: Visual Reasoning Tokens (\textit{vt}), which reside within the latent space of the DiT to capture low-level reasoning cues, and Textual Reasoning Tokens (\textit{tt}), which augment the text condition to capture high-level reasoning semantics. We subsequently analyze \textit{when, where, and how such tokens help}.

\subsection{Two Strategies of Reasoning Tokens}
\begin{figure}[t]
    \centering
    \includegraphics[width=\linewidth]{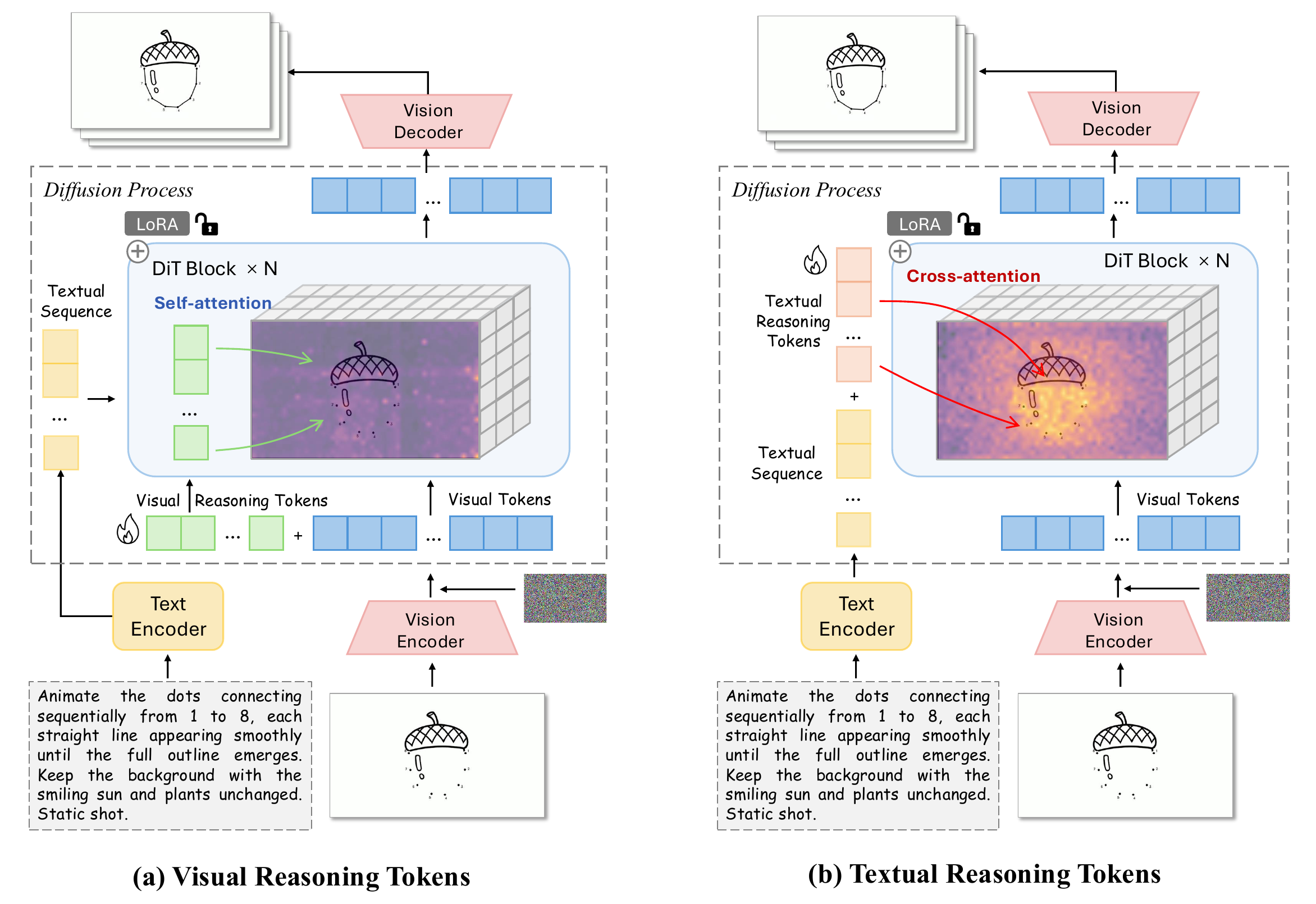}
    \caption{\textbf{Overview of reasoning-token designs.} (a) Visual Reasoning Tokens are inserted into the visual token sequence and interact with visual tokens through self-attention. (b) Textual Reasoning Tokens are appended to the text-conditioning sequence and guide generation through cross-attention.}
    \label{fig:reasoning_token}
\end{figure}

We now introduce our two reasoning-token strategies, as illustrated in \Cref{fig:reasoning_token}. The formulation is agnostic to the specific video generator and applies to any DiT-based architecture. For concreteness we instantiate it on Wan2.2-I2V-A14B, the same backbone used for \textsc{Wan-CoF}.

\paragraph{Visual reasoning tokens (\textit{vt}).}
A DiT video generator takes as input a sequence of visual tokens $x \in \mathbb{R}^{T \times H \times W \times d}$ that represent the compressed latent of the video, flattens them into a 1D sequence, and processes them through a stack of self-attention blocks. To host reasoning state directly in this visual space, we introduce $N_r$ learnable Visual Reasoning Tokens, denoted $r^v \in \mathbb{R}^{N_r \times d}$, initialized randomly and prepended to the visual token sequence before the first DiT block:
\begin{equation}
    z_0 = [r^v_1, r^v_2, \dots, r^v_{N_r}, x_1, x_2, \dots, x_M],
\end{equation}
where $M = T \times H \times W$ is the total number of visual tokens. Inside every DiT block, self-attention lets each visual token $x_i$ attend to every reasoning token $r^v_j$, and the reasoning tokens in turn attend to the full visual sequence and to each other. This bidirectional flow enables \textit{vt} to aggregate global context and distribute the resulting features back to the visual tokens. After the final DiT block, the reasoning tokens are discarded and the prediction is read from the visual portion of the output.

\paragraph{Textual reasoning tokens (\textit{tt}).}
A complementary place to host reasoning state is the text-conditioning side. The DiT consumes a text-conditioning sequence $c \in \mathbb{R}^{L \times d}$ produced by a text encoder, where $L$ is the length of that sequence. We introduce $N_t$ learnable Textual Reasoning Tokens, denoted $r^t \in \mathbb{R}^{N_t \times d}$, and prepend them to this sequence before it enters the cross-attention of each DiT block:
\begin{equation}
    c_0 = [r^t_1, r^t_2, \dots, r^t_{N_t}, c_1, c_2, \dots, c_L].
\end{equation}
The cross-attention in each block then lets every visual token $x_i$ attend to both the original text tokens and the new $r^t_j$, so $r^t$ supplies task-level information that supplements the per-prompt content of $c$. Unlike \textit{vt}, $r^t$ enters only as additional key/value context and never appears on the query side, so it is not refreshed by the visual sequence and is shared, unchanged, across every spatial patch and every temporal frame. This also means $r^t$ does not need to be removed before the final output readout. \textit{tt} therefore acts as a unidirectional, prompt-independent textual prior, complementary to the spatially grounded, bidirectional reasoning channel that \textit{vt} places inside the visual stream.


\subsection{Main Results}
We fine-tune Wan2.2-I2V-A14B with LoRA using \textit{vt} and \textit{tt} separately, yielding \textsc{Wan-CoF}$_{{vt}}$ and \textsc{Wan-CoF}$_{{tt}}$.
As shown in the Exploration of Reasoning Techniques rows of \Cref{tab:main_results,tab:genvire_results,tab:ruler_results,tab:viper_results}, both variants improve further over \textsc{Wan-CoF} on the headline metric of every benchmark. \textsc{Wan-CoF}$_{{vt}}$ reaches the highest aggregate score on Gen-ViRe and RULER-Bench, while \textsc{Wan-CoF}$_{{tt}}$ leads on MME-CoF and VIPER. At the sub-dimension level, the two variants specialize in opposite directions, in a way consistent with their design: \textit{vt} dominates dimensions that require maintaining a global, persistent plan inside the visual sequence, such as Planning in VIPER, Abstract, Analogy and Planning in Gen-ViRe, and Visual Stability in MME-CoF. In contrast, \textit{tt} dominates dimensions that benefit from a learnable, prompt-independent textual prior, such as Instruction Alignment in MME-CoF and Structural in VIPER.

\begin{figure}[t]
    \centering
    \begin{minipage}[t]{0.49\linewidth}
        \centering
        \includegraphics[width=\linewidth]{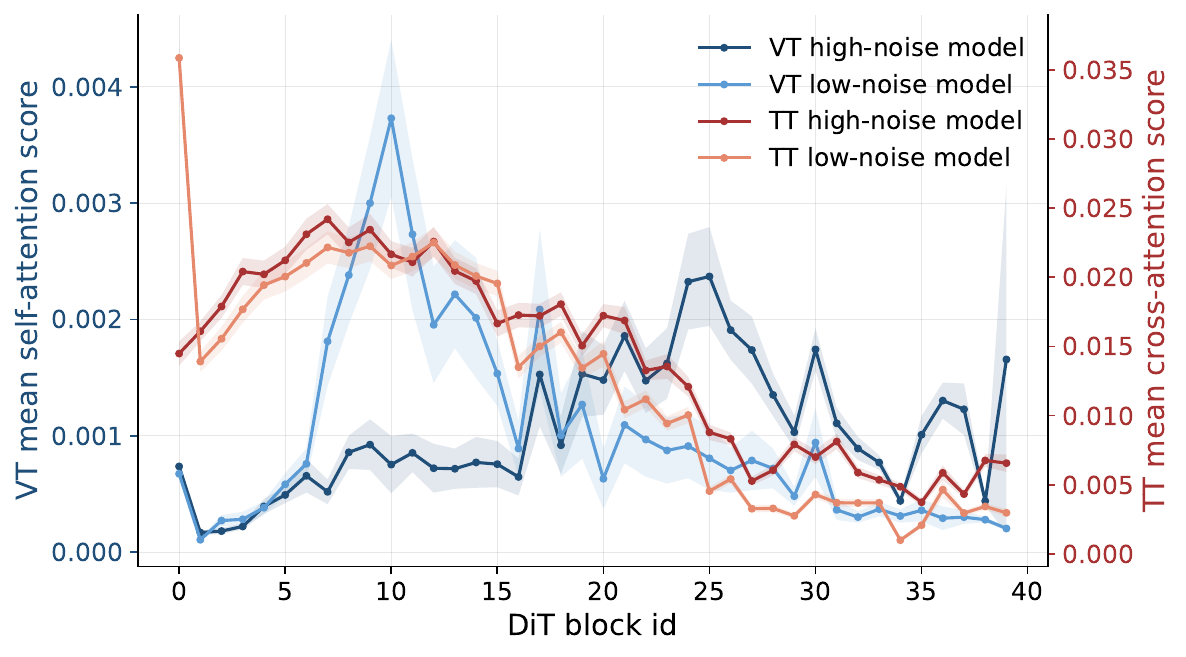}
        \centerline{\small (a) Depth-wise attention}
    \end{minipage}\hfill
    \begin{minipage}[t]{0.49\linewidth}
        \centering
        \includegraphics[width=\linewidth]{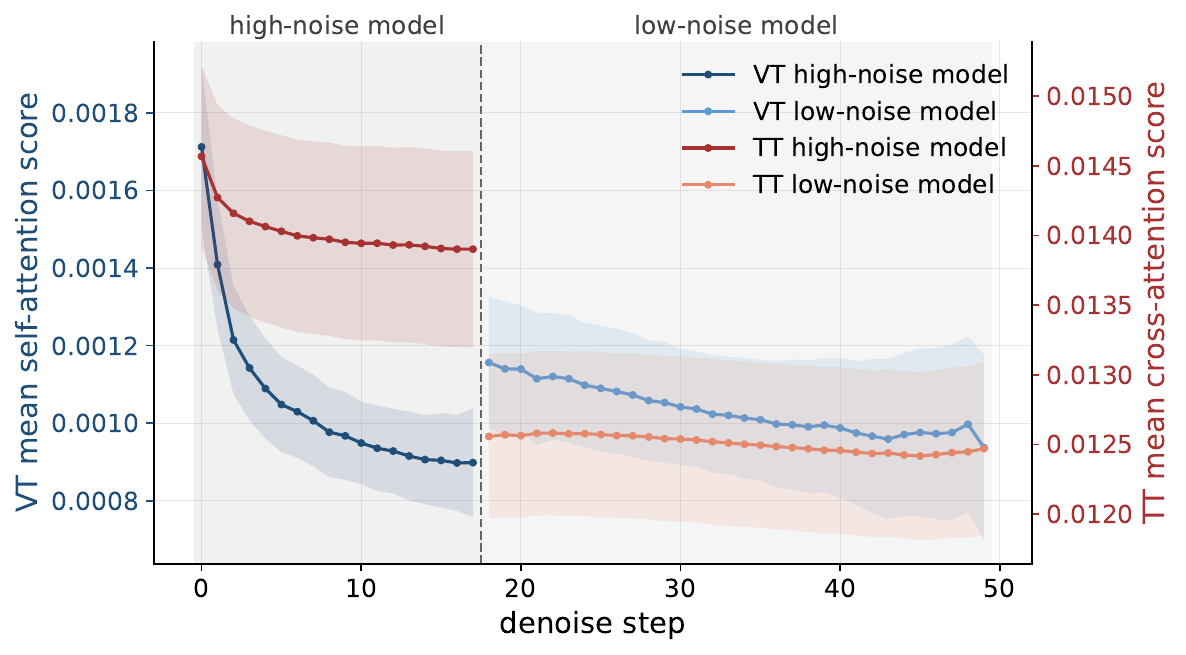}
        \centerline{\small (b) Denoising-step attention}
    \end{minipage}
    \caption{\textbf{Attention patterns of reasoning tokens.} We average attention over 120 MME-CoF cases during inference. For \textit{vt}, we report visual-token attention to \textit{vt} in self-attention; for \textit{tt}, we report visual-token attention to \textit{tt} in cross-attention. Panel (a) shows depth-wise attention, and panel (b) shows denoising-step attention.}
    \label{fig:reasoning_token_attention}
\end{figure}

\subsection{Attention Analysis}
To understand how the tokens operate, we extract attention during inference on 120 MME-CoF cases and average two signals: visual tokens' self-attention~\cite{vaswani2017attention} scores assigned to \textit{vt} and cross-attention~\cite{vaswani2017attention} scores assigned to \textit{tt}. We then inspect these attention patterns along four axes: DiT block depth, denoising step, spatial position within frames, and temporal position across frames.

\paragraph{Depth-wise attention.}
Panel (a) of \Cref{fig:reasoning_token_attention} shows that both \textit{vt} and \textit{tt} are used non-uniformly across DiT depth, suggesting that reasoning tokens mainly participate in specific computation stages rather than acting as constant global biases. \textit{vt} shows stronger layer selectivity: in the low-noise model it peaks around shallow-to-middle blocks, while in the high-noise model it remains active into later blocks and exhibits stronger mid-to-late spikes, consistent with visual reasoning tokens helping organize latent spatial-temporal structure when denoising is harder. \textit{tt} follows a smoother early-to-middle emphasis and then decays with depth, matching its role as a prompt-level prior supplied through cross-attention.

\paragraph{Denoising-step attention.}
Panel (b) of \Cref{fig:reasoning_token_attention} shows a similar pattern along the denoising trajectory. Both token types receive stronger attention early in denoising, especially in the high-noise model, where global semantics and motion structure are still underdetermined. \textit{vt} attention starts high and drops quickly in the high-noise model, indicating an early role in visual latent planning; in the low-noise model, it becomes flatter and lower, consistent with later refinement. \textit{tt} is smoother across denoising steps, but its high-noise curve stays above the low-noise curve, suggesting that task-level textual priors are most useful before the generation process settles.

\begin{figure}[t]
    \centering
    \includegraphics[width=\linewidth]{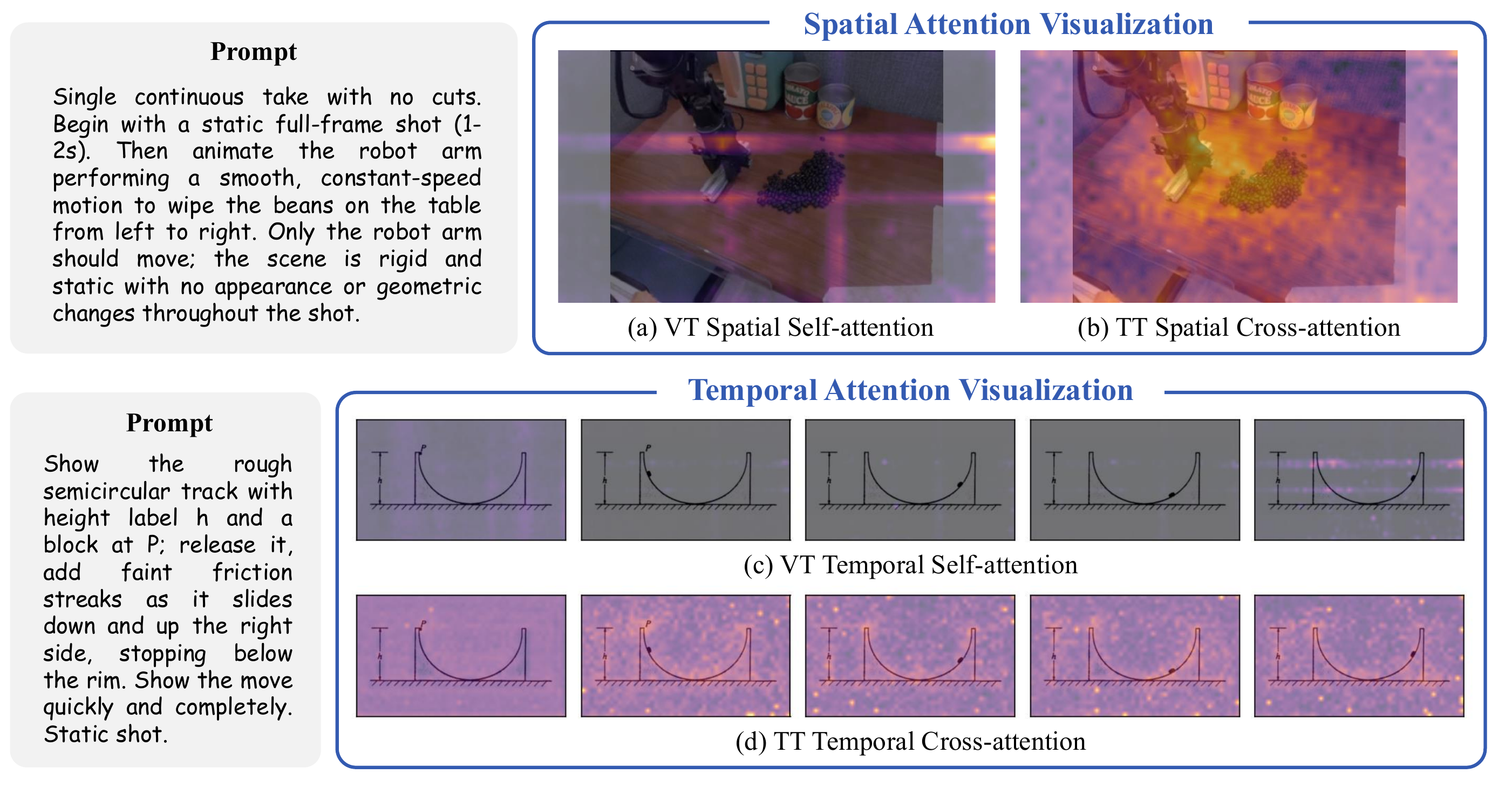}
    \caption{\textbf{Spatial and temporal attention visualizations.} Top: representative high-noise spatial attention maps for \textit{vt} self-attention and \textit{tt} cross-attention on an embodied manipulation prompt. Bottom: temporal attention maps for \textit{vt} and \textit{tt} on a physics-reasoning prompt, where frames are uniformly sampled over time and ordered from left to right.}
    \label{fig:attention_vis}
\end{figure}

\paragraph{Spatial attention.}
Within a single frame, RT attention reveals different spatial roles for \textit{vt} and \textit{tt}, as visualized in the top row of \Cref{fig:attention_vis}. \textit{vt} attention is relatively sparse and forms pronounced horizontal and vertical bands, with stronger responses near the gripper and the beans. This pattern suggests that \textit{vt} captures localized visual cues while retaining axis-aligned structure from the visual token grid, possibly influenced by Wan's separable 3D RoPE~\cite{su2024roformer}. In contrast, \textit{tt} attention is more diffuse and covers broader task-relevant regions, including the robot arm, table surface, and target objects, consistent with \textit{tt} acting as a high-level semantic prior rather than a localized spatial tracker.

\paragraph{Temporal attention.}
As shown in the bottom row of \Cref{fig:attention_vis}, \textit{vt} and \textit{tt} exhibit different temporal emphases rather than a simple object-tracking pattern. \textit{vt} attention is more visible at the beginning and end of the sequence, suggesting that visual reasoning tokens are involved when the model anchors the initial state and resolves the final consequence of the motion. \textit{tt} attention is less pronounced in the earliest frame, but becomes comparably salient across the later sampled frames, consistent with a prompt-level prior that remains active once the temporal context has been established. This contrast suggests a complementary temporal division: \textit{vt} emphasizes boundary states in the visual evolution, while \textit{tt} provides a steadier semantic constraint over the subsequent reasoning process.

%% file: sections/relatedwork.tex
\section{Related Work}

\subsection{Modern Video Generation Models}
Video generation models have advanced rapidly in visual quality, temporal consistency, and instruction following. Architecturally, the shift from U-Net-based diffusion~\cite{bar2024lumiere, ronneberger2015u} to DiT~\cite{peebles2023scalable} and flow-matching frameworks~\cite{ma2024sit} has enabled scalable training and higher-resolution synthesis. Open systems such as HunyuanVideo~\cite{kong2024hunyuanvideo}, HunyuanVideo-1.5~\cite{wu2025hunyuanvideo}, and Wan~\cite{wan2025} provide strong backbones for image-to-video and text-to-video generation, while proprietary systems including Kling~\cite{kuaishou2024kling}, Veo~\cite{GoogleDeepMind2025Veo3}, Seedance~\cite{seedance2026seedance}, and Sora~\cite{openai2025sora2} continue to push generation quality and controllability. These advances make video generation a promising substrate for visual reasoning, but high-fidelity generation alone does not guarantee coherent multi-step reasoning.

\subsection{Video Reasoning Benchmarks}
Reasoning has become a central capability for multimodal models~\cite{jiang2025mme,chen2025mint}, but conventional chain-of-thought pipelines~\cite{wei2022chain} often rely on text, static visual observations, external tools~\cite{hu2024visual}, or auxiliary image generation~\cite{li2025imagine}. Video generation provides another path, where reasoning unfolds through temporally connected frames~\cite{wiedemer2025video}. Recent benchmarks, including MME-CoF~\cite{guo2025video}, VideoThinkBench~\cite{tong2025thinking}, V-ReasonBench~\cite{luo2025v}, Gen-ViRe~\cite{liu2025can}, RULER-Bench~\cite{he2025ruler}, MMGR~\cite{cai2025mmgr}, TiViBench~\cite{chen2025tivibench}, and VIPER~\cite{li2025viper}, formalize and evaluate this capability from different perspectives. \textsc{OpenCoF} builds on these benchmarks but focuses on improving CoF reasoning through the \textsc{OpenCoF}-17K dataset and reasoning-token exploration.

\subsection{Video Generation Models for Reasoning}
Training video generation models for reasoning remains less explored than benchmarking. VR-Bench~\cite{yang2025reasoning} studies maze-solving and test-time scaling, Thinking in Frames~\cite{li2026thinking} focuses on domains such as mazes and tangram puzzles, and VBVR~\cite{wang2026very} scales video-reasoning data within a cognitive task suite. These efforts suggest that targeted supervision improves video reasoning, but they are often domain-specific or internally coupled between training and evaluation. \textsc{OpenCoF}-17K instead combines four data sources and supports transfer evaluation on independent external benchmarks, while \textsc{OpenCoF} further explores how dedicated reasoning techniques organize reasoning state through Visual and Textual Reasoning Tokens.

%% file: sections/conclusion.tex
\section{Conclusion and Limitations}
We present the \textsc{OpenCoF}-17K Dataset, a diverse video reasoning dataset for improving Chain-of-Frame reasoning in video generation models. Fine-tuning Wan2.2-I2V-A14B on the \textsc{OpenCoF}-17K Dataset yields \textsc{Wan-CoF}, showing that diverse temporal supervision improves CoF behavior across four external benchmarks. We further explore Visual and Textual Reasoning Tokens and analyze how they organize reasoning state through benchmark results and attention patterns. One limitation is that \textit{vt} and \textit{tt} are investigated separately. How to effectively combine them remains future work.

%% file: sections/appendix.tex
\subsection*{Overview}
We organize our supplementary material as follows.
\begin{itemize}
    \item Additional Implementation Details
    \begin{itemize}
        \item Backbone and LoRA Setup
        \item Model Variants
        \item Compared Systems
        \item Evaluation Protocol
    \end{itemize}
    \item Additional Ablation Study
    \begin{itemize}   
        \item Reasoning Token Count Ablation
    \end{itemize}
\end{itemize}

\section{Additional Implementation Details}
\label{sec:add_impl}

\subsection{Backbone and LoRA setup.}
Wan2.2-I2V-A14B adopts a Mixture-of-Experts~\cite{shazeer2017outrageously} denoising architecture with separate high-noise and low-noise experts across diffusion timesteps, an umT5~\cite{chung2023unimax} text encoder, a Wan-VAE for visual encoding and decoding, and a mainstream DiT denoiser. We fine-tune all three models (\textsc{Wan-CoF}, \textsc{Wan-CoF}$_{{vt}}$, and \textsc{Wan-CoF}$_{{tt}}$) from the same Wan2.2-I2V-A14B backbone using LoRA~\cite{hu2022lora}, applied only to the DiT denoiser while the umT5 text encoder and the Wan-VAE are kept frozen. Within each DiT block, LoRA targets the attention projections (\texttt{q}, \texttt{k}, \texttt{v}, \texttt{o}) together with the two MLP linear layers (\texttt{ffn.0}, \texttt{ffn.2}), with rank set to $32$. The learning rate is configured to $2 \times 10^{-5}$.

\subsection{Model variants.}
To attribute every gain to the dataset alone, we keep this architecture entirely unchanged and fine-tune it on \textsc{OpenCoF}-17K with LoRA, yielding \textsc{Wan-CoF}. For reasoning-token exploration, we run two independent fine-tuning experiments with LoRA on Wan2.2-I2V-A14B, one with only \textit{vt} and one with only \textit{tt}, yielding the models \textsc{Wan-CoF}$_{{vt}}$ and \textsc{Wan-CoF}$_{{tt}}$, where we set $N_r = N_t = 16$. \textsc{Wan-CoF} is trained on \textsc{OpenCoF}-17K for $2$ epochs, while \textsc{Wan-CoF}$_{{vt}}$ and \textsc{Wan-CoF}$_{{tt}}$ are trained for $5$ epochs to give the additional reasoning-token parameters more updates to converge.

\subsection{Compared systems.}
We compare \textsc{Wan-CoF} with both closed-source and open-source video generation models. The closed-source set covers state-of-the-art systems including Kling-v1~\cite{kuaishou2024kling}, the Seedance series~\cite{seedance2026seedance} Seedance-1.0-Pro and Seedance-1.5, the Veo-3~\cite{GoogleDeepMind2025Veo3} series Veo-3.0-Fast, Veo-3.0-Preview, Veo-3.0 and Veo-3.1, and Sora-2~\cite{openai2025sora2}. The open-source set includes HunyuanVideo-I2V~\cite{kong2024hunyuanvideo} and HunyuanVideo-1.5~\cite{wu2025hunyuanvideo}.

\subsection{Evaluation protocol.}
We evaluate our models on four video reasoning benchmarks: MME-CoF~\citep{guo2025video}, RULER-Bench~\citep{he2025ruler}, VIPER~\citep{li2025viper}, and Gen-ViRe~\citep{liu2025can}. Together, these benchmarks span reasoning categories well beyond the 11 task types in \textsc{OpenCoF}-17K. For RULER-Bench specifically, we restrict evaluation to the image-to-video cases, since our baseline does not support text-to-video generation. For every benchmark we follow its official judge model.


\begin{table*}[t]
  \centering
  \small
  \caption{\textbf{Ablation on the number of reasoning tokens.} Headline metrics on all four benchmarks for \textit{vt} and \textit{tt} with $n \in \{16, 32\}$; \textsc{Wan-CoF} is included as a reference. RULER Overall is the overall score on RULER-Bench, averaged across the four scoring dimensions (Instruction Following, Visual Consistency, Visual Fidelity, and Rule Coherence). Bold marks the better value within each $n{=}16$ / $n{=}32$ pair.}
  \label{tab:token_count_ablation}
  \begin{tabular}{l|cccc}
    \toprule
    \multirow{2}{*}{Model}
      & MME-CoF & Gen-ViRe & VIPER & RULER-Bench \\
      & Overall & Avg.     & POC$\uparrow$ & Overall \\
    \midrule
    \textsc{Wan-CoF}                                  & 1.30          & 0.391          & 7.5          & 56.8          \\
    \midrule
    \textbf{\textsc{Wan-CoF}$_{{vt}}$} ($n{=}16$) & \textbf{1.34} & \textbf{0.441} & 8.2          & \textbf{59.6} \\
    \textbf{\textsc{Wan-CoF}$_{{vt}}$} ($n{=}32$) & 1.28          & 0.416          & \textbf{8.9} & 58.5          \\
    \midrule
    \textbf{\textsc{Wan-CoF}$_{{tt}}$} ($n{=}16$) & \textbf{1.35} & 0.406          & \textbf{8.8} & \textbf{58.4} \\
    \textbf{\textsc{Wan-CoF}$_{{tt}}$} ($n{=}32$) & 1.34          & \textbf{0.412} & 6.8          & 57.7          \\
    \bottomrule
  \end{tabular}
\end{table*}


\section{Additional Ablation Study}
\label{sec:add_ablation}

\paragraph{Reasoning token count ablation.}
The exploration in \Cref{sec:exploration} fixes $N_r = N_t = 16$. To check whether a larger token budget changes our conclusions, we re-train \textsc{Wan-CoF}$_{{vt}}$ and \textsc{Wan-CoF}$_{{tt}}$ with $n = 32$ under otherwise identical settings and report the headline metric of every benchmark in \Cref{tab:token_count_ablation}. Doubling the token count does not produce a uniform improvement for either variant: each benchmark moves in a different direction, indicating that reasoning tokens do not act as generic capacity that scales with $n$. At the benchmark level, $n = 16$ outperforms $n = 32$ on three of the four benchmarks for both \textit{vt} and \textit{tt}, with $n = 32$ leading only on a single benchmark in each case. Combined with the fact that $n = 32$ also roughly doubles the reasoning-token parameters, we keep $n = 16$ as the default in the main text.